\relax
\documentclass[letterpaper]{article} 
\usepackage{spconf,amsmath,graphicx}
\usepackage{times}  
\usepackage{helvet}  
\usepackage{courier}  
\usepackage[hyphens]{url}  
\usepackage{graphicx} 
\usepackage[numbers]{natbib}  
\usepackage{caption} 
\DeclareCaptionStyle{ruled}{labelfont=normalfont,labelsep=colon,strut=off} 
\usepackage{algorithm}
\usepackage{algorithmic}
\usepackage{amsmath}
\usepackage{amssymb}
\usepackage{url}
\usepackage{newfloat}
\usepackage{listings}
\usepackage{hyperref}
\usepackage{pgfplots}
\pgfplotsset{compat=1.17}

\floatstyle{ruled}
\newfloat{listing}{tb}{lst}{}
\floatname{listing}{Listing}
\newcommand{\teasel}{\textsc{Teasel }}
\newcommand{\teaselns}{\textsc{Teasel}}

\title{\teaselns: A Transformer-Based Speech-Prefixed Language Model}
\name{Mehdi Arjmand \quad\qquad Mohammad Javad Dousti \quad\qquad Hadi Moradi}
\address{University of Tehran}

\usepackage{bibentry}

\begin{document}

\maketitle

\begin{abstract}
Multimodal language analysis is a burgeoning field of NLP that aims to simultaneously model a speaker's words, acoustical annotations, and facial expressions. In this area, lexicon features usually outperform other modalities because they are pre-trained on large corpora via Transformer-based models. Despite their strong performance, training a new \textit{self-supervised learning} (SSL) Transformer on any modality is not usually attainable due to insufficient data, which is the case in multimodal language learning. This work proposes a \textit{Transformer-Based Speech-Prefixed Language Model} called \teasel to approach the mentioned constraints without training a complete Transformer model. \teasel model includes speech modality as a dynamic prefix besides the textual modality compared to a conventional language model. This method exploits a conventional pre-trained language model as a cross-modal Transformer model. We evaluated \teasel for the multimodal sentiment analysis task defined by CMU-MOSI dataset. Extensive experiments show that our model outperforms unimodal baseline language models by 4\% and outperforms the current multimodal \textit{state-of-the-art} (SoTA) model by 1\% in F$1$-score. Additionally, our proposed method is 72\% smaller than the SoTA model.

\end{abstract}

\section{Introduction}

Human language is not limited to communicating with words; it employs acoustical annotations, body and head movements, and facial expressions besides using words to reinforce our intentions. For instance, people may strengthen their opposing viewpoints by emphasizing words or changing their vocal pitch. With the growth of social media data and the essential role of additional modalities in expressing an opinion, a new research area in \textit{Natural Language Processing} (NLP) called \textit{multimodal language analysis} has emerged \cite{challenge-hml-2020-grand}. This area includes multimodal sentiment analysis \cite{morency_towards_2011, soleymani2017survey} and multimodal emotion recognition \cite{bagher-zadeh-etal-2018-multimodal, busso2008iemocap}. 

In multimodal language analysis, different methods aim to combine heterogeneous sources of unimodal features. These methods leverage a spectrum of learning tasks, such as supervised \cite{tsai2019MULT}, multitask learning \cite{yu2021le}, reinforcement learnings \cite{chen2017multimodal}, and self-supervised \cite{khare2021self,yu2021le} approaches. However, in most cases, the textual modality outperforms other modalities; This superiority is mainly because the textual modalities are mainly from large language models pre-trained on huge corpora, while speech and visual features mainly use feature engineering-based approaches.

Recent large language models use \textit{Self-Supervised Learning} (SSL) and Transformer-based \cite{vaswani2017attention} methods utilizing large number of data points for pre-training in textual \cite{devlin2018bert,liu2019roberta, radford2019language, brown2020language}, speech\cite{baevski2020wav2vec}, vision \cite{dosovitskiy2020image} and multimodal visual-textual settings \cite{lu2019vilbert}. Collecting large amount of data is not easily attainable for most multimodal contexts, particularly multimodal language. Also, adding a new modality would require new data collection and SSL pre-training, which is not commonly feasible.

\begin{figure}[t]
\centering
\includegraphics[width=1\columnwidth]{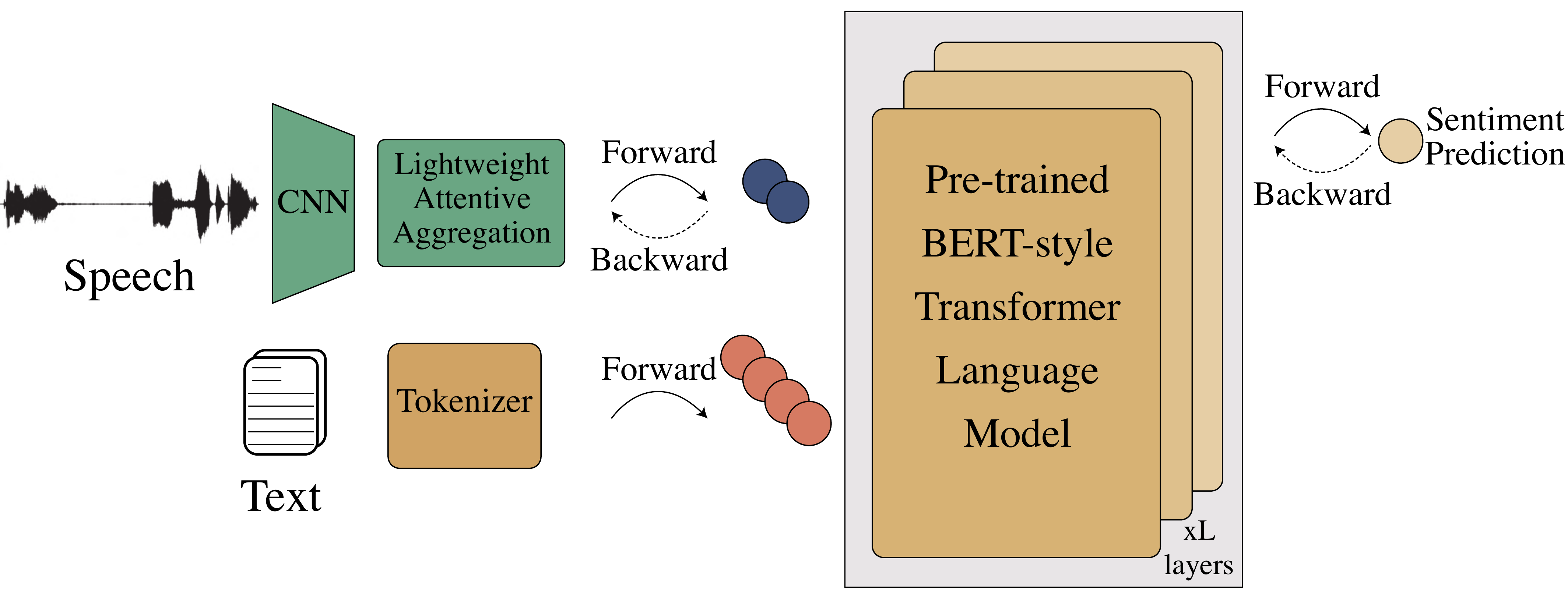} 
\caption{Process of fine-tuning \teasel model on multimodal sentiment analysis task. \teasel receives raw speech and text modalities as input and feed the speech prefixes and textual tokens to a pre-trained BERT-style language model. The $[CLS]$ output token will be used for fine-tuning.}
\label{fig:fine-tune}
\end{figure}

This work proposes a \textit{Transformer-Based Speech-Prefixed Language Model} (\teaselns) as an approach to generalize a pre-trained language model as a cross-modal attention module. \teasel introduces speech as a low-resource modality to a pre-trained Transformer-based language model without the need of pre-training the whole Transformer. In this approach, we pre-train a lightweight spatial encoder for speech to represent the speech modality as dynamic prefixes to a pre-trained language model and fine-tune this module besides the Transformer encoder layers in a downstream task. With this approach, \teasel model would exploit the pre-trained language model as a cross-modal multi-head attention during the pre-training and fine-tuning to a downstream task. 
\teasel does not require any data alignment. This is a key benefit in multimodal language analysis, where data alignment results in losing crucial intramodal information \cite{tsai2019MULT, yu2021le}.

We evaluate \teasel for the multimodal sentiment analysis task on CMU-MOSI \cite{zadeh2016mosi} dataset. \teasel shows 4\% percent improvement over the unimodal baseline and state-of-the-art (SoTA) performance gain of 1\% compared to the current multimodal SoTA using extensive experiments. \teasel does not need a fully-trained modality-specific Transformer and achieves better performance than late fusing of two fully-trained Transformers in a low-resource setting such as multimodal sentiment analysis. Moreover, \teasel uses approximately 72\% fewer parameters compared to late fusing approaches.

\section{Related Work}
\label{sec:related works}

\subsection{Human Multimodal Language Analysis}

Multimodal language analysis is a modern field of study in NLP which intends to model language interactions through different parallel simultaneous modalities. These multimodal studies typically contain text, audio, and visual features. Multimodal human language analysis applications include, but are not limited to, sentiment analysis \cite{zadeh2016mosi}, emotion recognition \cite{busso2008iemocap, zadeh2018multimodal}, humor detection \cite{hasan2019ur}, and sarcasm detection \cite{castro2019towards}.

\cite{baltruvsaitis2018multimodal} has defined representations, translation, alignment, fusion, and co-learning as five core challenges in multimodal machine learning. Among these challenges, \cite{baltruvsaitis2018multimodal} have described fusion as the concept of integrating information from multiple modalities to predict an outcome measure and classifies the fusion methods as \textit{early fusion}, \textit{hybrid fusions}, and \textit{late fusion}.

With the success of large Transformer-based language models, there are three common options to utilize these language models in multimodal language settings as described next.

First, certain multimodal language approaches freeze a Transformer and aim to fuse all modalities using tensor outer-products \cite{zadeh2017tensor,liu2018efficient}, Canonical Correlation Analysis based methods \cite{sun2020learning}, Attentive LSTM based methods \cite{zadeh2018multi, zadeh2018memory, chen2017multimodal, wang2019words, zadeh2018multimodal}, sequence to sequence based methods \cite{pham2019found}, cross-modal Transformer-based methods \cite{tsai2019MULT, zadeh2019factorized, hasan2021humor}, graph-based method \cite{yang2020mtgat}, and multi-task learning \cite{yu2021le}.

Second, as fine-tuning large pre-trained Transformer-based language models improve their performances on the downstream tasks significantly, recently, some approaches aim to employ this advantage in multimodal settings. \cite{rahman2020integrating} have proposed a method to fuse other modalities in the middle layer of a pre-trained Transformer-based language model in an aligned manner using a \textit{Multimodal Adaption Gate} (MAG) module. Later, with the popularity of Transformer-based models in Speech, \cite{siriwardhana2020jointly} has examined jointly fine-tuning lexicon and speech Transformer on the multimodal language task. They implemented \textit{Co-Attention} fusions and \textit{Shallow-Fusion} using an attentive and a straightforward late fusion of two BERT-style \cite{devlin2018bert} Transformers, respectively. 

Third, \cite{khare2021self} have proposed a BERT-style pre-training scheme for multimodal language settings, using text, speech, and visual features. They pre-trained their proposed model using one of the most extensive multimodal language datasets, consisting of around $1$ million video segments \cite{chung2018voxceleb2}. Nevertheless, this amount of data is much smaller than the textual corpora size and limits their model's performance.

Our method is a variation of the second approach, i.e., it fine-tunes a Transformer-based language model in a multimodal setting. In contrast to previous work, we utilize a pre-trained Transformer language model as cross-modal attention without needing an entirely different Transformer.

\subsection{Transformers and BERT-style Models}

A Transformer network is a model which utilizes the self-attention mechanism to encode sequential feature representations.
\cite{vaswani2017attention} have introduced the Transformer encoder-decoder network and attained a significant gain in Neural Machine Translations (NMT) task. 
Regarding this considerable improvement, \cite{devlin2018bert} have proposed \textit{Bidirectional Encoder Representations from Transformers} (BERT) model, which uses 12 layers of Transformer encoder. They pre-trained the model with \textit{Masked Language Model} (MaskedLM)  and \textit{Next Sentence Prediction} (NSP) tasks in a self-supervised manner. BERT has significantly improved after fine-tuning on the GLUE benchmark \cite{wang2018glue}, consisting of nine \textit{Natural Language Understanding} (NLU) tasks, including sentiment analysis and question answering. We refer the reader to \cite{devlin2018bert} for more details on the pre-training process. Moreover, \cite{liu2019roberta} introduced \textit{Robustly optimized BERT approach} (RoBERTa) model, which pre-trained BERT Transformer solely on MaskedLM task using larger corpora, and more effective training parameters. 

With regards to this notable success of BERT-style models in NLP, \cite{baevski2020wav2vec} introduced wav2vec 2.0, which contains a convolution layer as a temporal encoder, a quantization method, and a BERT-style Transformer as the backbone of the spatial encoder for speech encoding. They pre-trained the model on \textsc{LibriSpeech} dataset \cite{panayotov2015librispeech} containing 960 hours of speech audio. Subsequently, they achieved a significant enhancement on \textit{Automatic Speech Recognition} (ASR) task.

The key point of training these BERT-style models is the tremendous number of training data points. Accessing this massive number of data points is not always feasible for a multimodal language setting, and additionally, pre-training a new Transformer is a challenging task. This work employs a low-resource approach to include other modalities to a pre-trained BERT-style Transformer. Specifically, we use a method to introduce another modality as prefixes into a pre-trained BERT-style Transformer in the fine-tuning process. As such, there is no need to train a new task-specific Transformer.

\subsection{Generalization of Pre-Trained Language Model}

With the accomplishment of Transformers as language models on several textual tasks, \cite{lu2021pretrained} have examined the generalizations of fine-tuning these language models on various non-verbal modalities downstream tasks, such as image classifications and protein folding. They introduced \textit{Frozen Pre-trained Transformer} (FPT), using the GPT-2 \cite{radford2019language} language model as the backbone, and re-trained a small part of the Transformer layers. Their exhaustive set of evaluations demonstrate that it is feasible to achieve comparable results to a fully trained Transformer by only fine-tuning a small part of the pre-trained model on the downstream task.  Furthermore, following the success of the GPT-3 \cite{brown2020language} auto-regressive Transformer in few-shot learnings, \cite{tsimpoukelli2021multimodal} have introduced an image encoder that can generate prefix tokens for large frozen auto-regressive Transformer. They have achieved a method that performs better than baselines on several multimodal visual tasks, containing few-shot image classifications and the visual question answering. 

Inspired by \cite{tsimpoukelli2021multimodal} approach, we propose a method to train \textit{Lightweight Attentive Aggregation} (LAA) module, which generates speech prefixes for a pre-trained language model. However, in contrast to \cite{tsimpoukelli2021multimodal}, we fine-tune LAA module and the pre-tranined language model on the downstream task.

\section{Methodology}
\label{sec:methodolody}

\begin{figure}[t]
\centering
\includegraphics[width=1\columnwidth]{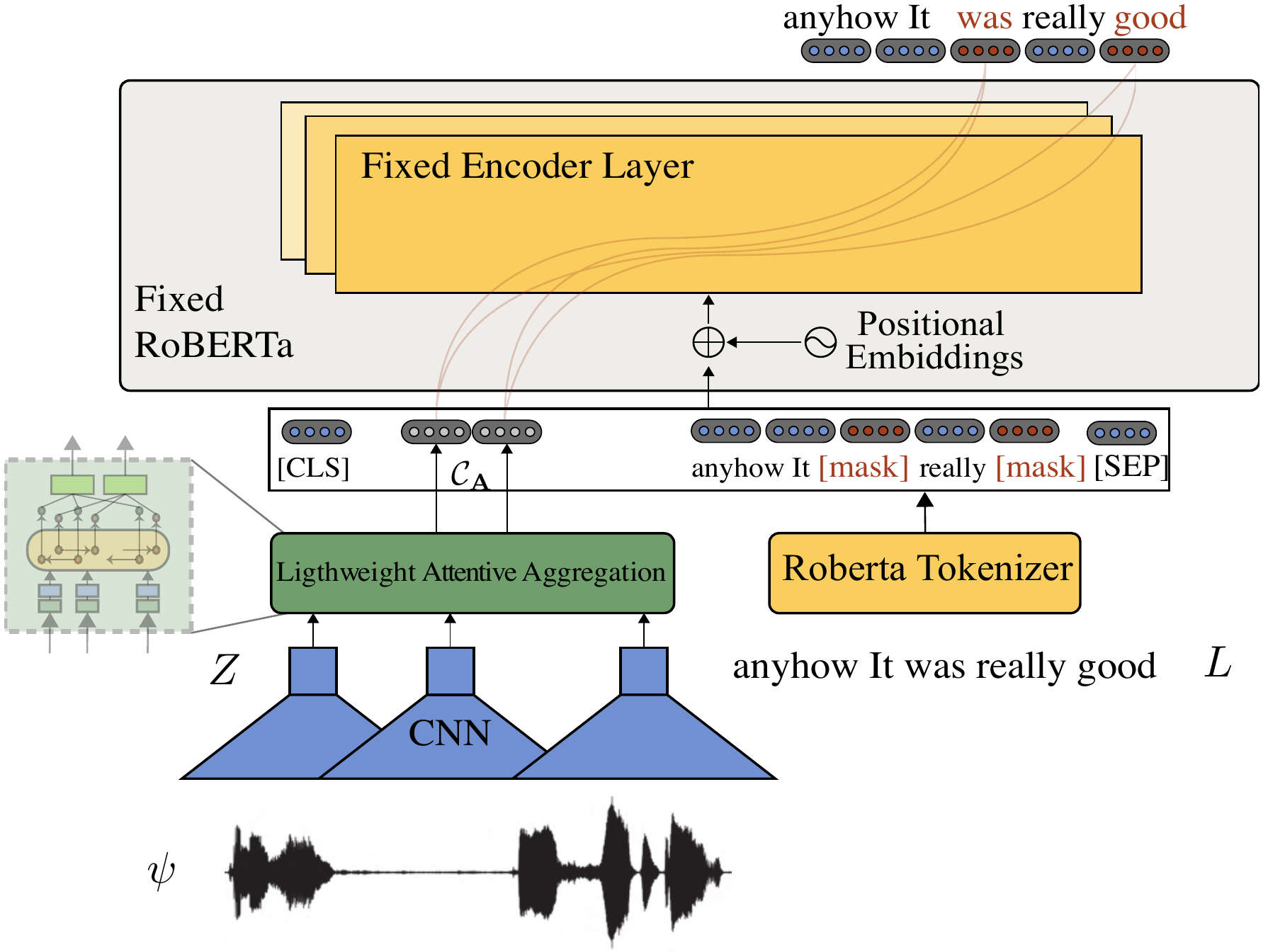} 
\caption{The architecture of \teasel during the pre-training process. \teasel takes $\psi$ as a raw speech and $L$ as corresponding text and calculates the MaskedLM task on the textual tokens to train LAA module.}
\label{fig:TEASEL-pre-training}
\end{figure}

This section describes our proposed \teasel model , \autoref{fig:TEASEL-pre-training}, that exploits conventional pre-trained language models as a cross-modal attention module. Fundamentally, \teasel model focus on representing speech features as prefixes to a pre-trained BERT-style language model (i.e., RoBERTa). We choose the speech modality as our additional modal among the speech and visual (facial expressions) modalities for three main reasons. First, speech datasets are more comprehensive. Second, working on raw speech data is less computationally intensive. Finally, previous works have shown that the speech modality serves more information in multimodal language settings \cite{tsai2019MULT, yu2021le}. 

Our method contains two steps, pre-training and fine-tuning. As for the pre-training step, we learn a representation to insert an additional modality to a fixed pre-trained language model using LAA module with relatively few training steps (8,000 total steps). Afterward, we fixed most part of LAA module and fine-tuned the pre-trained Transformer as a cross-modal Transformer on CMU-MOSI dataset for the multimodal sentiment analysis downstream task. We discuss the parts to fine-tune and parts to keep frozen in the ablation studies section.

In the rest of this section, we present the methodology of BERT-style Transformers, which serve as the backbone of our method. Respectively, we describe the speech feature extraction and LAA module. Finally, we describe \teasel model's pre-training and fine-tuning process.

\subsection{BERT-style Language Models}

We choose RoBERTa model \cite{liu2019roberta} for our conventional BERT-style encoder, because it has trained more robustly on much larger datasets. RoBERTa tokenizer decomposes sentence $L$  as 
\begin{equation}
    tokenizer(L) = \{[CLS], l_1, l_2, \dots, l_{T_{\mathbf{L}}}, [SEP]\}, 
\end{equation}
where $l_{i} \in \mathbb{R}^{d}$ represents a byte-level token, $T_{\mathbf{L}}$ denotes the number of time-steps in the textual modality, and $[CLS]$ and $[SEP]$ designate the beginning and the end of the sequence symbols, respectively. The standard published RoBERTa models consist of a base and a large model containing 12 and 24 layers, respectively, both of which are exclusively trained on the MaskedLM task \cite{devlin2018bert}. MaskedLM task intends to predict randomly masked tokens using the entire unmasked words of the sentence. MaskedLM loss masks 15\% of the tokens using a variety of masking methods. The standard MaskedLM \cite{devlin2018bert} uses the [MASK] token 80\% of the time, a random token 10\% of the time, and the unchanged token 10\% of the time; forcing the language model to predict the output token confidently. We use standard MaskedLM for our pre-training process. We refer the reader to \cite{devlin2018bert,liu2019roberta} for additional specifications of the pre-training process. 

\subsection{Speech Module}

As discussed, we need a unimodal pre-trained speech feature extractor model and then learn a relatively small spatial head at the top of representations to encode speech signals as a word token.

\subsubsection{Speech Temporal Encoder}

wav2vec 2.0 model \cite{baevski2020wav2vec} utilizes five layers of \textit{Convolutional Neural Networks} (CNN) as a temporal feature encoder and a BERT-style Transformer as the contextualized encoder. In this work, we select wav2vec 2.0 pre-trained fixed-parameter CNN as an audio feature encoder.

\begin{equation}
\label{eqn:CNN}
    \text{CNN}_{\theta_{\mathbf{W}}}(\psi) = \{z_1, z_2, \dots, z_{T_{\mathbb{A}}}; z_{i} \in \mathbb{R}^{d_{A}}\},
\end{equation}%
In \autoref{eqn:CNN} the CNN$_{\theta_{\mathbf{W}}}$ would take $\psi$ as a raw speech vector and represents $z_{i}$ as a speech latent feature for $T_{\mathbf{A}}$ time-steps, and  $\theta_{\mathbf{W}}$ denotes pre-trained parameters from wav2vec 2.0.

\subsubsection{Lightweight Attentive Aggregation (LAA)}

\begin{figure}[tp]
\centering
\includegraphics[width=1\columnwidth]{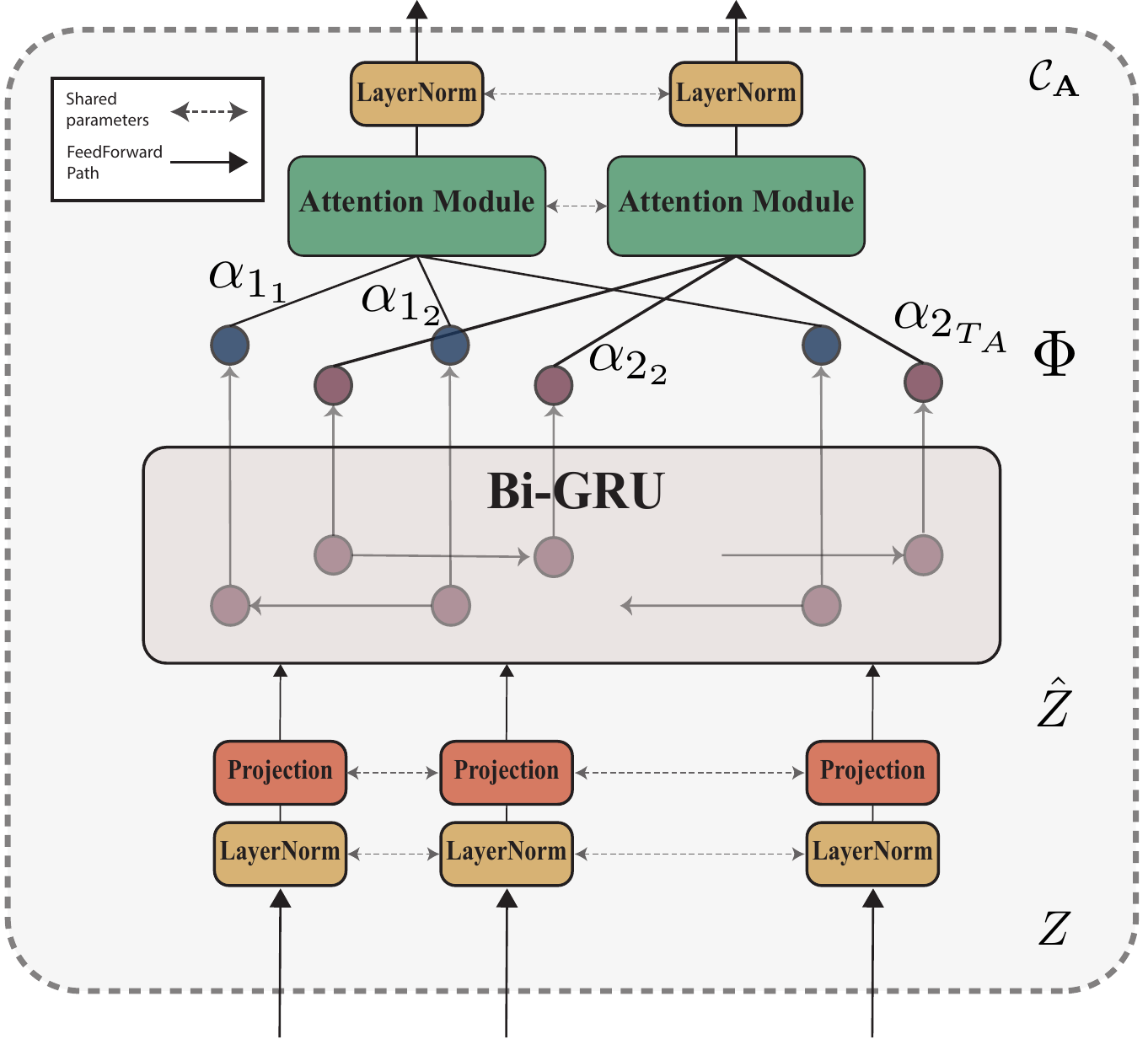} 
\caption{The architecture of Lightweight Attentive Aggregation (LAA) to aggregate the speech latent representation as prefixes with bidirectional attentive manner.}
\label{fig:aggregation}
\end{figure}

Although wav2vec 2.0 contains a well-defined Transformer network as a contextual encoder, for the sake of generalization, we train a small contextual encoder at the top of fixed features to simulate the effect of not having a fully trained Transformer for low-resources modalities.

Essentially, the speech LAA aims to encode $Z$ as prefix tokens for RoBERTa encoder. As \autoref{fig:aggregation} illustrates, this module performs a \textit{Bi-directional Gated Recurrent Unit} (BiGRU) at the top of speech fixed features to capture the information in a bidirectional behavior. More specifically, two output sequences of the BiGRU ($\Phi$) are defined as follows.

\begin{equation}
    \hat{Z} = LayerNorm(Z)
    \label{eqn:prj}
\end{equation}
\begin{equation}
    \Phi = BiGRU(W_{1} ^{\intercal} \hat{Z} + b_{1}),
\end{equation}
\begin{equation}
    \Phi = \{\{\phi_{\mathbf{1}, 1}, \ldots, \phi_{\mathbf{1}, T_{\mathbf{A}}}\}, \{\phi_{\mathbf{2}, 1}, \dots, \phi_{\mathbf{2}, T_{\mathbf{A}}} \}\},
\end{equation}%
where $W_{1} \in \mathbb{R}^{d_{\mathbf{A}} \times d_{\mathbf{A}}}$ and $b_{1} \in \mathbb{R}^{d_{\mathbf{A}}}$ respectively denotes projection weight and bias and $\Phi \in \mathbf{A}^{2 \times T_{\mathbf{A}} \times d_{\mathbf{A}}}$ denotes the BiGRU output sequences.

Next, we need an aggregation module to interpret the final prefixed tokens. Accordingly, we design LAA module based on the encoder used in the encoder-decoder models in the NMT field \cite{bahdanau2014neural}. Our LAA module calculates a dynamic weighted sum over $\Phi$ as,

\begin{equation}
        u_{k,i} = \sigma(W_{Agg_1}^{\intercal} \phi_{k,i} + b_{Agg_1}), k \in \{1, 2\}, 
\end{equation}


\begin{equation}
        \alpha_k = Softmax(W_{Agg_2}^{\intercal} u_k + b_{Agg_2}),  \alpha_{k} \in [0, 1],
\end{equation}

\begin{equation}
        \mathcal{C}_{\mathbf{A}} = \sum_{i=1}^{T} \alpha_{k,i} \phi_{k,i},
\end{equation}%
where $W_{Agg_{1}} \in \mathbb{R}^{d_{\mathbf{A}} \times d_{\mathbf{A}}}$, $b_{Agg_{1}} \in \mathbb{R}^{d_{\mathbf{A}}}$, $W_{Agg_{2}} \in \mathbb{R}^{d_{\mathbf{A}}}$, $b_{Agg_{2}} \in \mathbb{R}$, are feed forward's weights and biases, $\sigma$ is activation function, and $\mathcal{C}_{\mathbf{A}} \in \mathbb{R}^{2 \times d_{a}} $ are two RoBERTa-ready speech-prefixes. 

\subsection{Training Process}
\subsubsection{Pre-Training Process}

Similar to the training process mentioned in \cite{tsimpoukelli2021multimodal}, we used a fixed pre-trained language model as our Transformer encoder. We feed the sequence 
\begin{equation*}
    \{[CLS],\mathcal{C}_{\mathbf{A}}, l_1, l_2, \dots,[MASK], \dots, l_{T_{\mathbf{L}}}, [SEP]\}    
\end{equation*}%
to a pre-trained RoBERTa model. In the mentioned sequence, $[MASK]$ token is the MaskedLM masked token, and it only applies to lexicon tokens. We only calculate the loss function on the verbal output tokens, and its gradient only affects LAA module. In particular, the MaskedLM aims to predict $[MASK]$ tokens using other words and speech prefixes. In contrast to \cite{tsimpoukelli2021multimodal}, we used a BERT-style language model, absolute position embedding, and MaskedLM loss to train our LAA module. Moreover, in the ablation studies section, we study the importance of pre-training on the model's performance.

\subsubsection{Fine-tuning on Downstream Task}
As \autoref{fig:fine-tune} illustrates, we feed the speech prefixes and textual tokens to the pre-trained model and fine-tune it using an additional head at the top of the $[CLS]$ token. In contrast to \cite{tsimpoukelli2021multimodal} which Freezes their model on both pre-training and downstream tasks, we fine-tune a subset of our model besides the language model encoder on the downstream task. In the ablation studies section, we demonstrate the effect of freezing each part of the proposed model on the downstream task.  Furthermore, Algorithm \ref{alg:TEASEL_algorithm} shows a pseudocode for the whole proposed model. 

\begin{algorithm}[tb]
\caption{\teasel Model}
\label{alg:TEASEL_algorithm}
\textbf{Input}: speech $\psi$, sentences as $L$, \text{number of iterations} $epoch_{t}$, $epoch_{f}$, \text{learning rate}$\eta_{t}$, $\eta_{f}$\\
\textbf{Parameter}: Initialize $\theta_{LM}$ with pre-trained RoBERTa, $\theta_{\mathbf{W}}$ with pre-trained wav2vec 2.0, and $\theta_{\mathbf{A}}$, $\theta_{h}$ randomly \\
\begin{algorithmic}[1] 
\STATE $Z_{\mathbf{A}} \gets \text{CNN}_{\theta_{\mathbf{W}}}(\psi)$
\STATE $X_{\mathbf{L}} \gets tokenizer(L)$
\STATE // Pre-training process
\FOR{iteration $= 1,2, ..., epoch_{t}$}
    \STATE $\Phi_{\mathbf{A}} \gets BiGRU_{\theta_{\mathbf{A}}}(Z_{\mathbf{A}})$ // dropped Eq.~\ref{eqn:prj} for brevity
    \STATE $\alpha_{i} \gets Attention_{\theta_{\mathbf{A}}}(\phi_{i})$
    \STATE $\mathcal{C}_{\mathbf{A}} \gets \sum_{i=1}^{T} \alpha_{i} \phi_{i}$
    \STATE $[CLS; H_{A}; H_{L}] \gets RoBERTa_{\theta_{LM}}([\mathcal{C}_{\mathbf{A}}; L_{\mathbf{L}}])$
    \STATE $\nabla \theta_{\mathbf{A}} \gets \text{backpropagate}~~\mathcal{L}_{MLM}(H_{\mathbf{L}})$
    \STATE $\theta_{\mathbf{A}} \gets \theta_{\mathbf{A}} - \eta_{p} \nabla \theta_{\mathbf{A}}$ 
\ENDFOR
\STATE // Pre-training process
\STATE // Fine-tune $\theta_{LM}$ and $\theta_{\mathbf{A}}$ on the downstream task.
\FOR{iteration $= 1,2, ..., epoch_{g}$}
    \STATE $\Phi_{\mathbf{A}} \gets BiGRU_{\theta_{\mathbf{A}}}(Z_{\mathbf{A}})$
    \STATE $\alpha_{i, k} \gets Attention_{\theta_{\mathbf{A}}}(\phi_{i, k})$
    \STATE $\mathcal{C}_{\mathbf{A}} \gets \sum_{i=1}^{T} \alpha_{k, i} \phi_{k, i}$
    \STATE $[CLS; H_{A}; H_{L}] \gets RoBERTa_{\theta_{LM}}([\mathcal{C}_{\mathbf{A}}; L_{\mathbf{L}}])$
    \STATE $\Tilde{y} \gets ClassificationHead_{\theta_{h}}(CLS)$
    \STATE $\nabla \theta_{\mathbf{A}, LM, h}  \gets \text{backpropagate}~~\mathcal{L}_{MSE}(\Tilde{y}, y)$
    \STATE $\theta_{\mathbf{A}, LM, h} \gets \theta_{\mathbf{A}, LM, h}  -\eta_{f} \nabla \theta_{\mathbf{A}, LM, h} $ 
\ENDFOR
\STATE \textbf{return} $\theta_{\mathbf{A}, LM, h}$ as \teasel
\end{algorithmic}
\end{algorithm}

\section{Experimental Setup}
\label{sec:experimental setup}

\subsection{Dataset and Evaluation Methods}
We evaluated \teasel on CMU-MOSI \cite{zadeh2016mosi} for the multimodal sentiment analysis task.

\subsubsection{CMU-MOSI} 
\textit{Multimodal Opinion-level Sentiment Intensity dataset} (CMU-MOSI) \cite{zadeh2016mosi} contains 2199 video segments. Each has a sentiment intensity label with intensity in the range of $[-3, +3]$;  $+3$ denotes the solid positive sentiment, and $-3$ indicates the solid negative sentiment. To be consistent with prior work, we use the authors' published train/validation/test sets on CMU-Multimodal SDK \cite{zadeh2018multi}.



We followed the evaluation approach presented in \cite{tsai2019MULT} using binary accuracy and F$1$-scores on the non-zero human-annotated labels. More concretely, $[-3, 0)$ labels expresses negative and $(0, +3]$ predictions indicating positive sentiments. Additionally, we report the $7$-class accuracy (Acc$_{7}$), \textit{Mean Absolute Error} (MAE), and the correlation of the predicted labels with target labels to be consistent with the prior work \cite{zadeh2017tensor, liu2018efficient, tsai2018learning}.


\subsection{Baseline}
\label{ssec:baseline}

We consider a variety of SoTA unimodal and multimodal Transformers and fusion methods.
Experimental comparisons are reported in three sections:

First, we compare \teasel against unimodal (text only) large textual Transformers,

\begin{itemize}
    \item \textbf{RoBERTa}
    \cite{liu2019roberta} is a robustly pre-trained version of BERT Transformer. \cite{liu2019roberta} have published a base and large version of RoBERTa, using 12 and 24 layers, respectively. We used the base and large version of RoBERTa as a text-only baseline.
\end{itemize}

Second, we compare \teasel against popular multimodal fusion methods which use a Transformer as a feature extractor. These methods are as follows.
\begin{itemize}
    \item \textbf{Tensor Fusion Network (TFN)}
            utilizes tensor outer-product to combine three modalities in unimodal, bimodal, and trimodal fashion. Furthermore, it concatenates all six flattened vectors as a multimodal representation \cite{zadeh2017tensor} .
    \item \textbf{Low-rank Multimodal Fusion (LFM)} optimizes the TFN time complexity from exponential to linear using modality-specific low-rank factors \cite{liu2018efficient}.
    
    \item \textbf{Memory Fusion Network (MFM)} trains modality-specific generative factorized representations and joint multimodal features across all modalities for multimodal tasks \cite{tsai2018learning}.
    
    \item \textbf{Modality-Invariant and -Specific Representations for Multimodal Sentiment Analysis (MISA)} factorizes modalities into modality-invariant and modality-specific features and then fuses them to predict a label \cite{hazarika2020misa}.
    
    \item \textbf{Interaction Canonical Correlation Network (ICCN)} applies Deep CCA to dissolve hidden relationships between BERT sentence embeddings, audio sequence, and facial features \cite{sun2020learning}.

\end{itemize}

Finally, we compare \teasel performance against multimodal fusion Transformer-based methods.
\begin{itemize}
    \item \textbf{Multimodal Adaptation Gate-Transformer (MAG-Transformer)}
      is a module to generate a shift to the internal representation of Transformer models using visual and acoustic modalities. Combining the MAG module with a pre-trained Transformer would let them integrate speech and visual modalities to the middle level of the pre-trained lexicon Transformer \cite{rahman2020integrating}.
      
    \item \textbf{Multimodal Transformer (MulT)} performs directional cross-modal attention mechanism followed by self-attention for every two modalities to combine the heterogeneous multimodal signals using supervised settings \cite{tsai2019MULT}.
    
    \item \textbf{Self-Supervised Multi-task Multimodal (Self-MM)} employs multi-task learning and self-supervised learning to generate unimodal labels. Later, training the multimodal and unimodal representations to learn both variations and agreements among the modalities \cite{yu2021le}.
        
    \item \textbf{Shallow-Fusion} method jointly fine-tunes roberta-large and vq-wav2vec models on downstream tasks using a late fusion method \cite{siriwardhana2020jointly} .
\end{itemize}

\subsection{Implementation Details}
In this subsection, we review the implementation specification of both pre-training and fine-tuning steps.  We implemented \teasel using PyTorch \cite{paszke2019pytorch} and HuggingFace frameworks \cite{wolf2019huggingface} with a fixed random seed to guarantee reproducibility. We applied grid search for hyperparameter tuning during fine-tuning with learning rates in range of \{1e-4, 5e-5, 2e-5, 1e-5\}, LAA dropout in the range of \{0.0, 0.05, 0.10, 0.2, 0.3, 0.4, 0.5\}, batch sizes in the range of \{8, 16, 32\}, and warm-up steps using \{0\%, 10\%, 20\%, 30\%, 40\%\} of total step. We used the validation set of CMU-MOSI to find the best hyperparameters. The best setting was 2e-5 as learning rate, 0.1 as LAA dropout, and warmup step of 10\% of total fine-tuning steps. Also, we used a batch size of 32 and 16 during pre-training and fine-tuning, respectively. We pre-trained and fine-tuned \teasel using AdamW optimizer \cite{loshchilov2017decoupled} , GELU activation function \cite{hendrycks2016gaussian}, cosine scheduler with afrosaid warmup steps. We pre-trained \teasel using 100 hours of \textsc{LibriSpeech} in 8,000 cumulative steps ($9$ epochs) and fine-tuned \teasel on CMU-MOSI with $3$ epochs.

\section{Discussion of Experimental Results}
\label{sec:results}
\subsection{Quantitative Analysis}

\begin{table}[t]
\caption{\label{tab:MOSI}Results for multimodal sentiment analysis on CMU-MOSI. (\textit{B}) and (\textit{R}) indicate sentence representations are from BERT and RoBERTa, respectively.} 
\resizebox{\columnwidth}{!}{
\begin{tabular}{ l ||c c c c c}
\hline
\multicolumn{6}{|c|}{CMU-MOSI}                                    \\ \hline
                   & $\text{MAE}^{l}$ & $\text{Corr}^{h}$ & $\text{Acc}_{2}^{h}$  & $\text{F}1^{h}$ & $\text{Acc}_{7}^{h}$  \\ \hline 
TFN\textsuperscript{$\dagger$}(\textit{B})         & 0.901  & 0.698  & 80.82  & 80.77 & 34.94  \\ 
LMF\textsuperscript{$\dagger$}(\textit{B})          & 0.917  & 0.695  & 82.47  & 82.47 & 33.23  \\ 
MFM\textsuperscript{$\dagger$}(\textit{B})          & 0.877  & 0.706  & 81.72  & 81.64 & 35.42  \\ 
ICCN(\textit{B})          & 0.860  & 0.710  & 83.00   & 83.00  & 39.00   \\ \hline
MulT                        & 0.871  & 0.698  & 83.00   & 82.80  & 40.00   \\ 
BERT\textsuperscript{$\ddagger$}               & 0.739  & 0.782  & 85.20   & 85.20  & -      \\ 
Self-MM(\textit{B})       & 0.713  & 0.798  & 85.98   & 85.95  & -   \\  
MAG-BERT(\textit{B})     & 0.739  & 0.796  & 86.10   & 86.00  & -      \\ 
MAG-XLNet                & 0.675  & 0.821  & 87.90   & 87.90  & -      \\ 
Shallow-Fusion(\textit{R})             & \textbf{0.577}  & -       & 88.27  & 88.57 & \textbf{48.92} \\ \hline
\textit{roberta-base}\textsuperscript{$\ddagger\ddagger$}                & 0.704  & 0.807 & 85.31    & 85.37  & 46.36  \\ 
\textit{roberta-large}\textsuperscript{$\ddagger\ddagger$}            & 0.687  & 0.835 & 86.89    & 86.91  & 46.65 \\ \hline \hline
\teaselns\textsuperscript{$\dagger\dagger$}(\textit{R}) (ours)           & 0.644  & \textbf{0.842}   & \textbf{89.33}  & \textbf{89.31} & 47.52  \\ \hline

\multicolumn{4}{l}{$^h$\footnotesize{indicates higher is better.}}\\
\multicolumn{4}{l}{$^l$\footnotesize{indicates lower is better.}}\\
\multicolumn{4}{l}{$^\dagger$\footnotesize{Models are proposed by \cite{hazarika2020misa}.}}\\
\multicolumn{4}{l}{$^\ddagger$\footnotesize{Models are suggested by \cite{rahman2020integrating}}.} \\
\multicolumn{4}{l}{$^{\dagger\dagger}$\footnotesize{Models are produced using the same environment.}}

\end{tabular}}

\end{table}

In \autoref{tab:MOSI}, we present the results of experiments on CMU-MOSI dataset on several benchmarks. Benchmark results are from their corresponding paper discussed in the Baseline section, except TFN, LMF, and MFM models, which \cite{hazarika2020misa} reproduced using BERT sentence representations as a textual modality.

First, we evaluate \teasel against the text-only popular Transformers. \teasel has performed better than text-only RoBERTa base and large models. The p-value of the student t-test between \teasel and RoBERTa base is $p < 0.05$, demonstrating significant improvement over RoBERTa base model.
    
Second, in our experiments, we significantly improve over methods that apply the frozen Transformer-based features, principally because fine-tuning pre-trained Transformers for the downstream task would significantly improve the performance of Transformers. 
    
Eventually, we examine \teasel against networks which fine-tune a Transformer for the downstream task. Our method has outperformed Self-MM \cite{yu2021le}, MAG-BERT, MAG-XLNet \cite{rahman2020integrating}, and Shallow-fusion \cite{siriwardhana2020jointly} methods in most metrics. Unlike MAG-Transformer methods, \teasel does not require an aligned feature and explicitly feeds speech representations to the Transformer. As for Shallow Fusion, \teasel outperforms their best setup in F$1$-score, Acc$_2$ metrics, but their method has a superior result for the Acc$_7$ and MAE utilizing a fully trained speech BERT-style Transformer. Moreover, \teasel uses $133.4$M parameters while Shallow-Fusion uses approximately $483.6$M parameters, which indicated \teasel uses 72.4\% fewer parameters.

                   


\begin{figure*}[ht]
\centering
\includegraphics[width=0.8\textwidth]{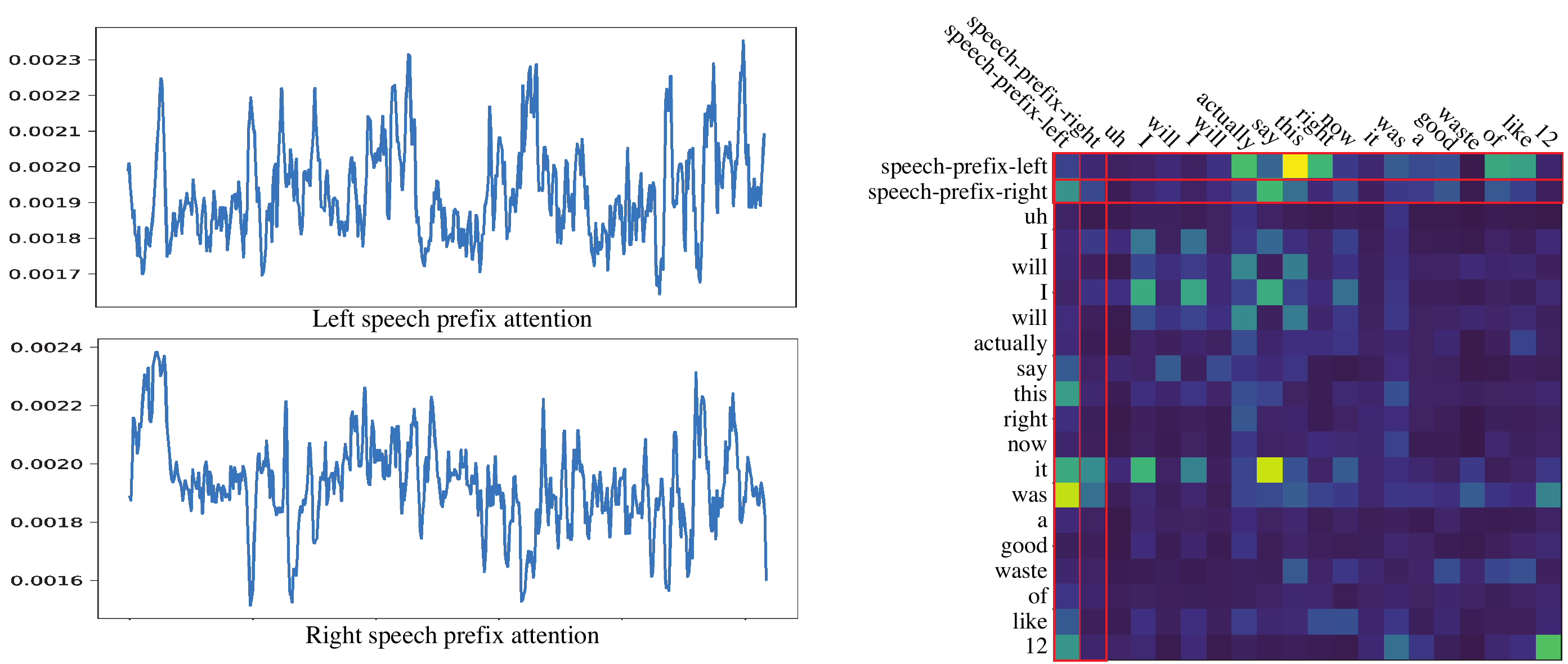}
\caption{The attention activation layer of a random data point in the test set. Left plots are LAA  speech attention values and the right figure demonstrates the attention activation layer in the attention encoder. As the cross-modal attention demonstrates, different word tokens are attending to speech prefixes.}
\label{fig:visualization}
\end{figure*}

\subsection{Ablation Studies}
\label{ssec:ablation studies}

We study the following subjects to examine the usefulness of some decisions in designing \teasel.

\subsubsection{Question 1: Does pre-training effects performance of \teasel?}

\begin{figure}[hpt]
\centering
\begin{tikzpicture}[scale=0.8]
\begin{axis}[
    xlabel={Pre-Training Step},
    ylabel={Performance on CMU-MOSI},
    xmin=-500, xmax=8500,
    ymin=0.84, ymax=0.91,
    xtick={0,2000,4000,6000,8000},
    ytick={0.84, 0.85, 0.86, 0.87, 0.88, 0.89, 0.90, 0.91},
    legend pos=north west,
    ymajorgrids=true,
    grid style=dashed,
]


\addplot[
    color=blue,
    mark=square,
    ]
    coordinates {
    (0,0.8577)(2000,0.8843)(4000,0.8798)(6000,0.8892)(8000,0.8894)
    };
\addlegendentry{Valid F$1$-Score}



\end{axis}

\end{tikzpicture}
\caption{\label{fig:pre-training-step} Results of fine-tuning different checkpoints of pre-trained \teasel using the same condition. Performance of each fine-tuned model on validation set is evaluated. For each checkpoint, F1-score of different pre-trained checkpoints is reported.}
\end{figure}
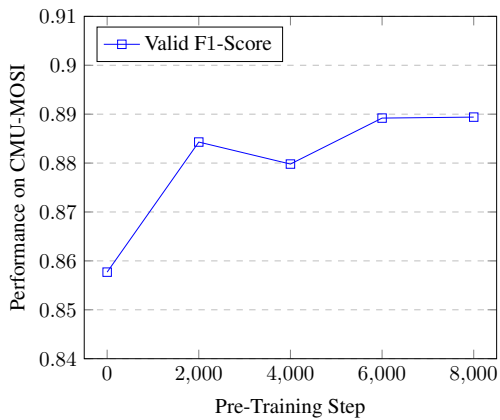

We examined the effect of pre-training process on the multimodal sentiment analysis downstream task. We trained LAA module using 100 hours of \textsc{LibriSpeech} dataset (around 28k data points) for 8,000 training steps. Subsequently, we saved the model every 2,000 steps and fine-tuned the saved checkpoints for the multimodal sentiment analysis on CMU-MOSI dataset. As \autoref{fig:pre-training-step} demonstrates, pre-training LAA module improves \teaselns.


\subsubsection{Question 2: Does the whole model needs fine-tuning?}

We trained the new classification head along with some parts of pre-trained \teasel model. We have done exhaustive experiments on which parts of LAA modules should be fixed for the fine-tuning part. As \autoref{tab:sub-fin-tuning} represents, we obtained the best result on solely fine-tuning RoBERTa encoder and LAA besides the classification head. We Believe this is primarily because CMU-MOSI does not contain many data points (2,199), and fine-tuning BiGRU and Projection hurt the model's performance significantly.

\begin{table}[t]
\caption{\label{tab:sub-fin-tuning} Results of fixing or fine-tuning different subsections of Lightweight Attentive Aggregation. + in each row indicates that we fine-tuned that module and all of the above ones together. For instance, in the second row we fine-tuned roberta-encoder and Attention module and keep the rest parts fixed. Also, We understand fine-tuning the embedding layer would hurt the model's performance. }
\resizebox{\columnwidth}{!}{
\begin{tabular}{l ||c c c c c}
\hline
\multicolumn{6}{|c|}{Frozen or Fine-tuning Parts of \teasel}                                    \\ \hline
                   & $\text{MAE}^{l}$ & $\text{Corr}^{h}$ & $\text{Acc}_{2}^{h}$  & $\text{F}1^{h}$ & $\text{Acc}_{7}^{h}$  \\ \hline 
robert-encoder            & 0.6861  & 0.8228  & 87.80  & 87.80 & 47.81 \\ 
+ Attention            & \textbf{0.6443}  & \textbf{0.8423}  & \textbf{89.33}  & \textbf{89.31} & 47.52  \\ 
+ BiGRU            & 0.6531  & 0.8334  & 88.41  & 88.40 & \textbf{48.98}  \\ 
+ Projection               & 0.7142  & 0.8192  & 85.37   & 85.28  & 47.08   \\ 
All Parameters           & 0.6556  & 0.8360  & 88.57   & 88.57  & 48.69   \\ \hline
roberta-base  & 0.7042 & 0.8072 & 85.31  & 85.37 & 46.36  \\ 
roberta-large  & 0.6870 & 0.8347 & 86.89  & 86.91 & 46.65 \\\hline

\end{tabular}}
\end{table}

\subsubsection{Question 3: Why we need an aggregation for the speech prefixes tokens?}

The Transformer's attention complexity is $O(n^2)$, where $n$ is the length of a sequence \cite{devlin2018bert}. Letting the pre-trained Transformer use additional sequences might improve the entire process, but that would significantly increase the computation complexity. As we aggregate speech features before feeding them to a pre-trained Transformer, this would not increase the Transformer's complexity quadratically. Moreover, our attention module uses linear computation complexity with respect to sequence length in order to attend BiGRU's output sequences. We illustrate and discuss the effectiveness of speech prefixes in the qualitative analysis section.


\subsection{Qualitative Analysis}

We illustrate \teaselns's random attention activation layer to investigate whether the fine-tuned Transformer encoder has learned to attend the speech prefixes. \autoref{fig:visualization} exhibits a random datapoint from CMU-MOSI test set, "lXPQBPVc5Cw$\_$16", in which the speaker says "\textit{uh I will I will actually say this right now it was a good waste of like 12 (dollars)}" with a non-negative voice tone, which caused the label to be weakly positive (i.e., $+0.2$). In \autoref{fig:visualization}, RoBERTa attention masks demonstrate that different word tokens attend to speech prefixes that are from an attentive module. While in BERT-style Transformers, there is excellent attention between each token and $[CLS]/[SEP]$ tokens, we removed the $[CLS]$ and $[SEP]$ tokens for better illustration.

\section{Conclusion}
\label{sec:conclusion}

We proposed a Transformer-Based Speech-Prefixed Language Model called \teasel to model multimodal sentiment analysis without training a new cross-modal Transformer. We implemented a Lightweight Attentive Aggregation module to create an efficient spatial encoding, which creates speech prefixes for a pre-trained Transformer. \teasel successfully utilized a pre-trained Transformer as a cross-modal attention module. Comprehensive analyses validated that \teasel outperforms Transformer based text-only language model by 4\% and achieves an approximately 1\% superior result than the current multimodal SoTA. Extensive experiments verify the effectiveness of various steps of \teasel model. We hope this work can provide a new perspective on pre-trained Transformers' generalization for multimodal language environments.

\bibliographystyle{IEEEbib}
\bibliography{teasel}

\end{document}